\pdfoutput=1
\documentclass[11pt]{article}

\usepackage{acl}

\usepackage{times}
\usepackage{latexsym}

\usepackage[T1]{fontenc}
\usepackage[utf8]{inputenc}

\usepackage{microtype}
\usepackage{subcaption}
\usepackage{graphicx}
\usepackage{float}

\title{KULCQ: An Unsupervised Keyword-based Utterance Level Clustering Quality Metric}

\author{Pranav Guruprasad \\  Georgia Institute of Technology \\ \texttt{pranavguruprasad0@gmail.com} \And
  Negar Mokhberian \\ University of Southern California \\ \texttt{nmokhber@usc.edu}
  \AND
  Nikhil Varghese \\ Got It AI \\ \texttt{nikhil@bot-it.ai} \And
  Chandra Khatri \\ Got It AI \\ \texttt{chandra@bot-it.ai} \And
  Amol Kelkar \\ Got It AI \\ \texttt{amol@bot-it.ai}}

\begin{document}
\maketitle
\begin{abstract}
Intent discovery is crucial for both building new conversational agents and improving existing ones. While several approaches have been proposed for intent discovery, most rely on clustering to group similar utterances together. Traditional evaluation of these utterance clusters requires intent labels for each utterance, limiting scalability. Although some clustering quality metrics exist that do not require labeled data, they focus solely on cluster geometry while ignoring the linguistic nuances present in conversational transcripts. In this paper, we introduce \textbf{K}eyword-based \textbf{U}tterance \textbf{L}evel \textbf{C}lustering \textbf{Q}uality \textbf{(KULCQ)}, an unsupervised metric that leverages keyword analysis to evaluate clustering quality. We demonstrate KULCQ's effectiveness by comparing it with existing unsupervised clustering metrics and validate its performance through comprehensive ablation studies. Our results show that KULCQ better captures semantic relationships in conversational data while maintaining consistency with geometric clustering principles.
\end{abstract}

\section{Introduction}
Over the past few years, there has been a proliferation of frameworks to develop task-oriented dialog systems. Most such systems utilize intent classification to discover the intent of a user. Since intents are specific to the domain, they vary significantly depending on the tasks that they are configured for. Thus, it is crucial to perform open intent discovery \cite{Zhang_Xu_Lin_Lyu_2021} in order to build task-oriented dialog systems. While there are metrics like Normalized Mutual Information (NMI) \cite{nmi} and Adjusted Rand Index(ARI) to evaluate the performance of open intent discovery, they require annotated intent labels. This is hard to scale and slows down the process of building task-oriented dialog systems. There is a need to measure the goodness of the intent clusters obtained from open intent discovery in an unsupervised way.
% todo explain measures of clustering say that silhouette is very common

%The Silhouette coefficient \cite{rousseeuw1987silhouettes} is one of the most popular metrics that measures cluster quality in an unsupervised manner. The Silhouette score of a cluster answers the question - ``are the ‘within’ (intra-cluster) dissimilarities small compared to the ‘between’ (inter-cluster) dissimilarities?" (more details in Section \ref{sec:sil}).
Clustering textual data is used in various fields such as Computational Social Science \cite{compsocsci}, Conversational AI \cite{Cuayhuitl2019DeepRL}, and Recommendation Systems \cite{7019655}. % todo for each of these mentioned fields cite other works
Intent detection is one of the most crucial tasks in conversational AI, implemented via classification or clustering. Classification being supervised, requires labels, making clustering a natural choice for discovery and scalablility. Various unsupervised clustering evaluation metrics such as the Silhouette coefficient \cite{rousseeuw1987silhouettes}, Calinski-Harabasz index \cite{chindex}, Davies-Bouldin index \cite{dbindex}, and Dunn index \cite{dunnindex} take in data belonging to any domain and apply the same algorithm each time, as long as there is a similarity measure defined between pairs of data points. These metrics only focus on the geometry of the clusters, and are agnostic to the domain-specific nuances of the data. Therefore, it is important to have metrics that evaluate clusters by leveraging aspects that are important to the domain the data belongs to.

In this paper we introduce a \textbf{K}eyword-based \textbf{U}tterance \textbf{L}evel \textbf{C}lustering \textbf{Q}uality \textbf{(KULCQ)} measure which is specific to analysing clusters of conversation data (Section \ref{sec:kulcq}). The motivation behind leveraging keywords for conversational data is to capture similar intents across utterances despite differences in sentence formation and helping words used - which may lead to a big difference in the sentence embeddings. For example: "How can I get a taxi from A to B?", and "I want a taxi from A to B, can you help?", may differ in representation even though taxi, A, and B are the most important words in both cases.
 
In our experiments we focus on evaluating the clustering of intents from user query utterances in dialogues between customer support agents and customers. We show in Section \ref{sec:badbad} that our metric can be used as a universal clustering metric, because it follows the trend of a standard metric in an adverse scenario, and correctly captures the general quality of the clusters. In the Sections \ref{sec:badgood} and \ref{sec:highgenvar} of the paper, we highlight the shortcomings of a standard domain-agnostic metric such as Silhouette coefficient in cases specific to conversational data. Our proposed metric handles these cases and provides an advantage in accurately measuring the clustering quality of conversational data.
% todo different ways for intent detection: clustering and classification
% say that classification is supervised, so clustering is better
% which works do clustering cite them 
% 

% evaluation metrics for conversational AI 
% which works use clustering in conv ai
% say that we don't want to change representation, but sometimes representations are considering wrong because of some extra details in the utterance 
% motivate why we only use in conversation, "How can I get a taxi from A to B?", "I want a taxi from A to B, can you please help me?"

% Many algorithms can be used to perform topic modeling, but one very common one is Latent Dirichlet Allocation (LDA). LDA is a generative probabilistic model that assumes that each document is made up of a distribution of a fixed number of topics and each topic is made up of a distribution of words. A big challenge when trying to use LDA (and many other topic modeling algorithms) is deciding how many topics to actually use, which is a necessary model hyperparameter. Obviously, if that is what we’re hoping to get out of the analysis then this is a problem.

% Evaluating performance, knowing the ground truth labels
% In this case, we happen to also know the ground truth labels so we can see how well our loss function correlates with performance. We can manually inspect how well the models did on some of the ground truth clusters:
\section{Methodology}
This section contains an overview of the Silhouette method: a well-known, unsupervised metric for evaluating clusters of data of any domain (Section \ref{sec:sil}). We compare it with our algorithm, as it is inspired by the Silhouette method and is based off of it. We then define our measure KULCQ (Section \ref{sec:kulcq}), which is an enhanced metric specific for evaluating clusters of conversational data.

\subsection{Silhouette}
\label{sec:sil}
For a given object x belonging to cluster y, the Silhouette coefficient \cite{rousseeuw1987silhouettes} is defined as a combination of its intra-cluster metric and inter-cluster metric. The intra-cluster measure is defined as $a(i) = $ average dissimilarity of x to all other objects of cluster y. The inter-cluster metric is defined as: 
\setlength{\abovedisplayskip}{4pt}
\setlength{\belowdisplayskip}{4pt}
 $$b(x) = min_{i\in C, i\neq y} d(x, i)$$
in which $C=\{1,2, ..., N\}$ is the set of all clusters and $d( x, i) $ is defined as: $d( x, i) =$ average dissimilarity of object x to all objects of cluster i. Finally, $a(x)$ and $b(x)$ are aggregated as: 
\setlength{\abovedisplayskip}{4pt}
\setlength{\belowdisplayskip}{4pt}
$$silhouette(x)=\frac{b(x)-a(x)}{max\{a(x),b(x)\}}$$
\begin{figure*}[ht!]
% \hspace{1em}
\begin{subfigure}{.48\textwidth}
  \centering
  % include first image
  \includegraphics[width=\linewidth]{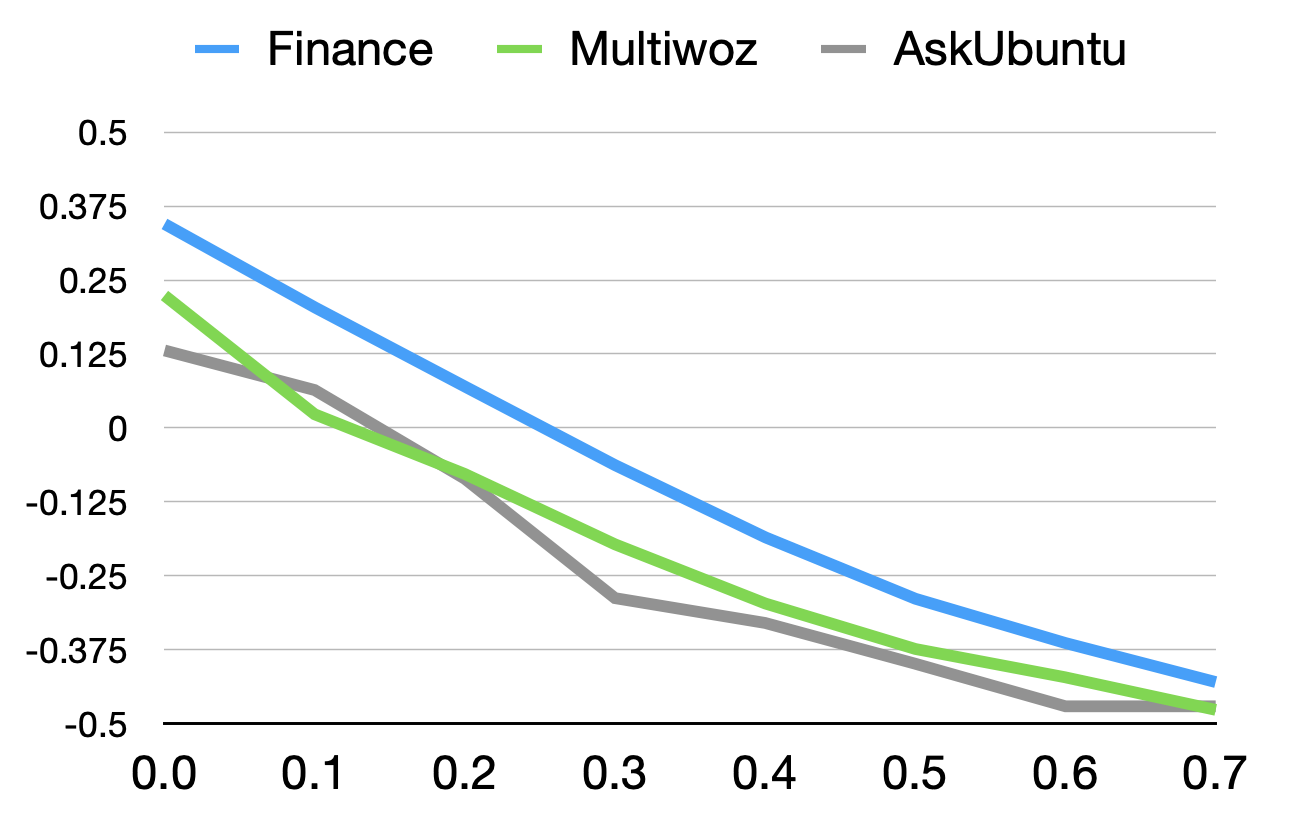}  
  \caption{KULCQ}
  \label{fig:sub-first}
\end{subfigure}
\hspace{1em}
\begin{subfigure}{.48\textwidth}
  \centering
  % include second image
  \includegraphics[width=\linewidth]{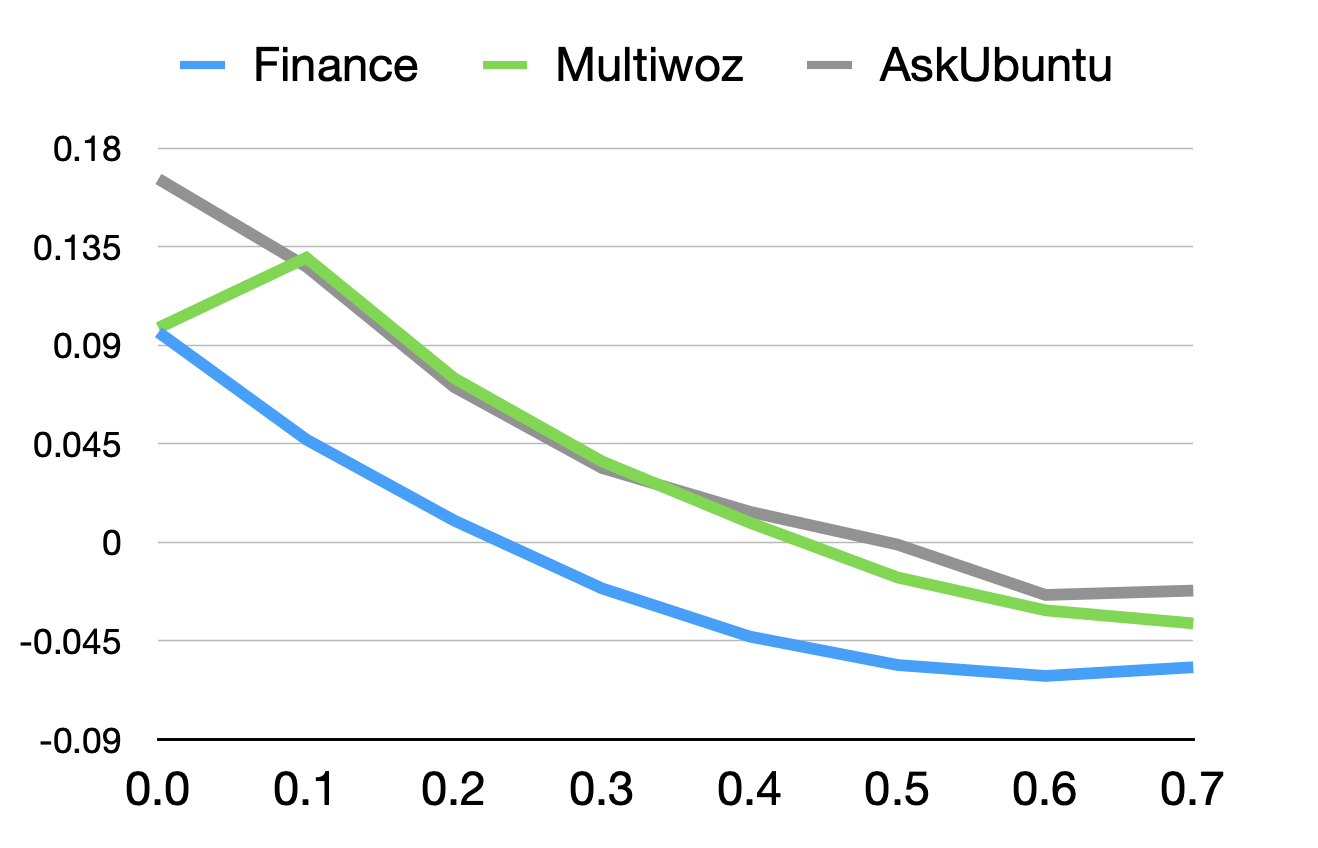}  
  \caption{Silhouette}
  \label{fig:sub-second}
\end{subfigure}
% \hspace{1em}
\caption{X-axis shows the probability of perturbing each utterance's label, and the Y-axis denotes the clustering metric. We observe that both the KULCQ and Silhouette scores decrease when the probability of perturbation increases. The decrease in KULCQ score is more monotonic compared to that of the Silhouette score. }
\label{fig:corrupt}
\vspace{-0.5cm}
\end{figure*}
\subsection{KULCQ}
\label{sec:kulcq}
Similar to the Silhouette method, the KULCQ score is a combination of an intra-cluster measure, and an inter-cluster measure. Let us say we have a set of $N$ clusters $C=\{1,2, ..., N\}$. Each cluster $i$ includes the set of utterances $U_i$. First, we extract keywords from each utterance. The keywords from each utterance are a combination of keywords extracted from the utterance using two libraries - KeyBERT\footnote{\url{https://github.com/MaartenGr/KeyBERT}} and Yake\cite{yake}. KeyBERT defines keywords as the words that are most similar to the document as a whole in terms of embeddings, the similarity metric being cosine similarity. Yake on the other hand uses statistical features to identify and rank the most important keywords. Thus, we use a combination of these two methods to get the keywords from each utterance. A keyword can either be a unigram or bigram. % todo the mathematical definition of keywords (what is considered as a keyword?) procedure of finding keywords should be completely described here
Based on the utterance keywords over the entire cluster, we find the top $n$ most frequent keywords in each cluster ($K(cluster_i)$ denotes the top $n$ most frequent keywords for cluster $i$, $i \in C$). A centroid for each cluster $i$ ($c_i$) is calculated as the weighted average of its utterances:
$$c_i = \sum_{j \in U_i}{w_i^j \times R(u_j)},$$
in which $U_i$ is the set of utterances belonging to cluster $i$ and $R(u_j)$ is the representation of $u_j$ using an embedding method.
Weight of utterance $j$ for calculating the centroid of the cluster $i$ is the proportion of the top $n$ frequent keywords appearing in the utterance. More precisely:
$$w_i^j = \frac{K(cluster_i) \cap K(utt_j)}{|K(cluster_i)|}.$$

The intra-cluster score is calculated for the cluster as a whole, which is then assigned as the intra-cluster score to each of the utterances in the cluster. The intra-cluster metric is defined as the average of of the distances from each utterance to the centroid which was calculated for the cluster.
$$a(x) = \frac{1}{|U_i|}\sum_{j \in U_i}{D(R(u_j),R(c_i))},$$
in which $D$ is a distance function (we use cosine distance in our experiments). The inter-cluster metric for utterance $x$ which belongs to cluster $y$ is defined as the weighted average of distances of utterance $x$ to the centroids of other clusters:
$$b(x) =\sum_{i \in C, i\neq y} {{w'_i}^y \times D(R(x),R(c_i))}.$$
We calculate ${w'_i}^y$ as the reciprocal of the number of overlapping top $n$ frequent keywords between the clusters $i$ and $y$,
$${w'_i}^y=\frac{1}{K(cluster_i) \cap K(cluster_y)}.$$

By using the reciprocal of the number of overlapping keywords between clusters as a weight while calculating the inter-cluster distance, we discourage different clusters from having a similar set of the most important keywords, and encourage different clusters to be far away from each other - in terms of the most important keywords of each cluster. Finally, the KULCQ score for an utterance is calculated by combining the inter- and intra-cluster scores of the utterance, similar to Silhouette:
$$KULCQ(x) = \frac{b(x)-a(x)}{max\{a(x),b(x)\}}$$ %This translates to an increased inter-cluster score for the utterance - if the cluster it belongs to and the other cluster have lesser important keywords in common. 

\section{Experiments and Results}
\paragraph{\textbf{Datasets.}} We use three popular, public human-annotated datasets, thus containing gold intent labels. Detailed information about the datasets can be found in appendix section \ref{sec:datasets}. We hypothesize that KULCQ is useful for any text clustering task. Given our application being conversational, we limit our experiments to related datasets.

Further in this section we analyze the behavior of the Silhouette and KULCQ metrics on clustered conversational data, and showcase using a few scenarios what KULCQ brings to the table in a conversational setting. The experimental setup we used can be seen in appendix section \ref{app:setup}.

\subsection{Noise injection}
\label{sec:badbad}

% adding noise
We conducted this experiment to test the correctness of KULCQ. We observe the values given by the Silhouette and KULCQ metrics when different amounts of noise are injected into the clusters. Noise here is defined as the perturbation to a cluster by taking utterances from its correctly assigned cluster and putting it into a different(wrong) cluster, and vice versa. Given the annotated clusters, we perturb the cluster of each utterance with probability $p$. In figure \ref{fig:corrupt} we observe that both the KULCQ and Silhouette scores drop as we inject more noise. This validates the behavior of our metric, and proves that it follows the trends of standard metrics. The result of this experiment also validates that our proposed metric reflects the noise in the clustering. Moreover, we can see in figure \ref{fig:corrupt}, that the KULCQ metric is more sensitive to noise injection in a cluster. The drop in KULCQ values are much more as compared to the drop in Silhouette values when the probability of perturbing each utterance's label is increased.

\subsection{Bad Geometry, Good Semantics}
\label{sec:badgood}
\begin{figure}[ht!]
    \centering
    \includegraphics[width=\columnwidth]{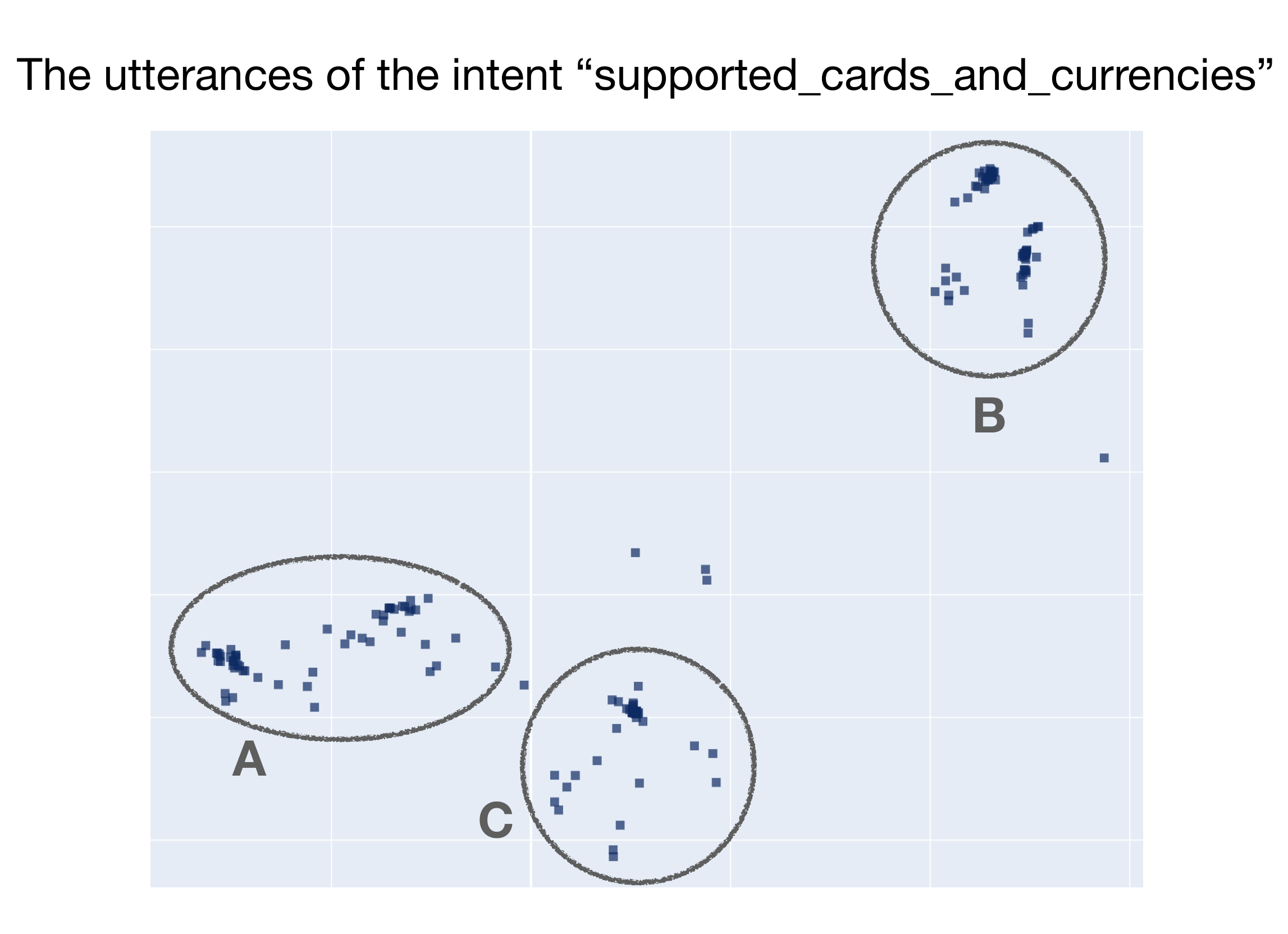}
    \caption{The cluster "supported cards and currencies" receives a low score from the Silhouette metric, but the KULCQ metric evaluates it as a fairly good cluster. The source of confusion for the Silhouette score is the vast difference in representation of the three subgroups of utterances. Region A includes the utterances that talk about credit card, region B includes the utterances that talk about currencies, and region C includes the utterances that mention a specific bank's name.}
    \label{fig:finance}
\vspace{-0.3cm}
\end{figure}
Here we depict the scenario where having a clustering evaluation metric specific to conversational or text data such as KULCQ is more helpful compared to a domain agnostic metric such as the Silhouette score. We specifically focus on clusters whose utterances due to minor differences in the words present, end up far from each other in the embedding space. In figure \ref{fig:finance} we show the cluster ``supported cards and currencies" from \textit{Finance} dataset as a TSNE~\cite{tsne} plot. At first glance it does not seem to be a great cluster, because its utterances have been divided into three regions. The Silhouette coefficient reflects this immediate conclusion, and gives us a low average Silhouette score for utterances of this cluster, based purely on the geometry of the cluster. The Silhouette score for this cluster is $-0.05$ which is the second lowest score among the $77$ clusters of the \textit{Finance} dataset. However, based on the KULCQ scores, this cluster ranks as the $26th$ highest. KULCQ's judgement turns out to be more reasonable as we look at the cluster and its utterances more closely. We observe that \textit{region A} includes utterances that talk purely about credit cards (\textit{e.g. ``What US cards do you accept?", ``Do you accept US credit cards?"}), \textit{region B} includes the utterances that talk purely about currencies (\textit{e.g. ``What currencies can I use?", ``Do you have a list of the cards and currencies supported?"}) and \textit{region C}'s most important keyword is a specific bank's name (\textit{e.g. ``Is American Express supported for adding funds?", ``How do I use American Express to top up my account?"}). The annotators of the dataset have assigned all such utterances to the same cluster because despite the minor differences, the intent of the utterances in the cluster remain the same. KULCQ, by leveraging the importance of keywords in representing the intent of the utterances, is able to gain information beyond what meets the eye at first glance - the geometry of the cluster.

%\subsection{Good Geometry, Good Semantics}
%\label{sec:goodgood}
% this will help to show that our measure is universal 
% Our method is universal

\subsection{Penalizing high generality and variance}
\label{sec:highgenvar}
We also evaluate a scenario where the Silhouette metric does not penalize certain kinds of clusters enough for containing highly varied utterances, which can in turn lead to difficulties in conversational settings. Contrarily, KULCQ metric results in a bad score for the same clusters, thus providing an advantage in conversational AI problems.

One example can be seen in the \textit{AskUbuntu} dataset, for the cluster - "Software Recommendation". The Silhouette metric gives this cluster a small positive score of 0.04, whereas the KULCQ metric gives this a score of -0.01. Clearly the KULCQ metric penalizes this cluster much more severely as compared to the Silhouette metric. As seen in Table \ref{tab:examples}, the utterances of this cluster are widely varied, and it is highly unlikely that these utterances will be clustered together by any automatic intent discovery algorithm. Moreover, in many conversational AI problems, it may not be a wise idea to cluster utterances that are as varied as the ones in Table \ref{tab:examples} into one cluster with a general intent. Consider an application of conversational AI such as a chatbot: a slight change in intent can require a completely different action by the bot. In scenarios such as this, it is important that the utterances are clustered in more specific intents as opposed to a highly general one. Keeping this in mind, it is important that clusters such as the "Software Recommendation" cluster are penalized significantly in a conversational setting. By leveraging keywords, KULCQ is able to reward tightly-knit, specific clusters, and penalize the generic and varied ones, which can offer a great advantage for conversational AI applications.

Additionally, we performed clustering over all utterances across datasets using HDBScan\cite{hdbscan} and K-means\cite{kmeans} and observed similar patterns depicted in Figure \ref{fig:corrupt} and Figure \ref{fig:finance}. Since the clusters obtained aren't manually labeled, we conclude our findings solely based on the ground truth intent labels.

% TODO we should here also show that: Always the cluster None has the lowest Silhouette and KULCQ score. However this does not happen right now! we should be able to explain why this does not always happen with kulcq

\section{Conclusion}

We define a new metric (KULCQ) for evaluating the clustering quality of text data, specifically conversational data. We compare KULCQ closely with a well-known, unsupervised cluster evaluation metric - the Silhouette coefficient, and show some of the advantages KULCQ provides in a conversational setting when compared to a domain-agnostic metric which relies solely on the geometry of the cluster. We also depict the correctness of our algorithm and prove that it can be used as a universal clustering metric for text data.

We realize that using the reciprocal of an integer such as the number of overlapping keywords as a weight directly may lead to significant fluctuations in the scale of the overall metric, even for small differences in the integer value, which in turn may not reflect the quality of clustering accurately. We are investigating other ways to incorporate the amount of keyword overlap between clusters, such as - 1) using the reciprocal of a probability measure instead, and 2) considering the average cosine similarity between overlapping keywords. We aim to share the findings in the future work.

\bibliography{anthology,custom}
\bibliographystyle{acl_natbib}
\appendix
\section{Datasets Description}
\label{sec:datasets}
We used public conversational datasets for our experiments, in which all the utterances are annotated with their respective intent labels by humans. Table \ref{tab:dataset} contains a detailed description of all the datasets used.

\begin{table}[ht!]
\begin{tabular}{|p{2cm}|p{5cm}|}
\hline
\textbf{Dataset Name} & \textbf{Description}\\
\hline
\hline
\textbf{Finance} \cite{casanueva-etal-2020-efficient} &  This dataset includes online banking queries annotated with their corresponding intents. Consists of 10003 utterances and 77 intents.   \\
\hline
\textbf{MultiWOZ} \cite{budzianowski-etal-2018-multiwoz} & Multi-Domain Wizard-of-Oz dataset consists of human-human written conversations related to booking hotels, restaurants, and other topics. \\
\hline
\textbf{AskUbuntu} \cite{braun-EtAl:2017:SIGDIAL} & The dataset includes 162 questions and answers from AskUbuntu website\footnote{\url{https://askubuntu.com}}. The utterances are annotated with five intent labels. \\
\hline
\end{tabular}
\caption{The datasets used for our experiments.}
\label{tab:dataset}
\end{table}

\section{Experimental Setup}
\label{app:setup}
For calculating the KULCQ score (definition in section \ref{sec:kulcq}) we extract keywords from each utterance. We used Python libraries KeyBERT and Yake for extracting the keywords.

For sentence-level embeddings we use the \textit{``all-MiniLM-L6-v2"} model from the \textit{Sentence Transformers (sbert)} repository \cite{reimers-gurevych-2019-sentence}. 

The KULCQ and Silhouette scores were calculated for all the datasets mentioned above on an utterance level, cluster level, and dataset level. For both metrics - the KULCQ and Silhouette scores, a cluster-level score is an average of the utterance-level scores for that cluster, and dataset-level score is an average of cluster-level scores over all the clusters in the dataset. 
This was done in order to compare and showcase the advantage KULCQ has to offer over another standard, unsupervised, utterance-level clustering quality evaluation metric.

Note that the actual intent names/labels are not used anywhere, and only the utterances per intent are treated as clusters, from which the keywords are extracted.

\section{Example Utterances for the Software Recommendation cluster in the AskUbuntu dataset}
\label{sec:soft_recom}
\begin{table}[ht!]
\begin{tabular}{|l|}
\hline
Utterances \\
\hline
\hline
Is there an SSH connection manager?       \\
\hline
MySQL GUI Tools      \\
\hline
What developer text editors are available for Ubuntu? \\
\hline
Is there an application for reading mobi files? \\
\hline
Can you recommend a password generator? \\
\hline
\end{tabular}
\caption{Example utterances from the 'Software Recommendation' cluster in the AskUbuntu dataset}
\label{tab:examples}
\end{table}
% This is an appendix.

\end{document}